\documentclass[letterpaper,10pt,conference]{ieeeconf}
\IEEEoverridecommandlockouts  %

\usepackage{iftex}
\usepackage[fleqn]{amsmath}
\ifXeTeX
  \usepackage{newtxtext,newtxmath}  %
\else
  \usepackage{amssymb}
\fi
\usepackage{graphicx}
\usepackage{xcolor}
\usepackage{url}
\urldef{\projecturl}\url{https://omron-sinicx.github.io/point-hand/}
\usepackage{booktabs}
\usepackage{dblfloatfix}
\usepackage{multirow}
\usepackage{diagbox}
\definecolor{hcbBg}{HTML}{F2D5C8}    %
\definecolor{hcsBg}{HTML}{E8E6DC}    %

\newcommand{\method}{PointHand}
\newcommand{\pvr}{keypoint hand visual representation}

\newcommand{\kdp}{DP-KP}
\newcommand{\condBase}{\textsc{Baseline}}
\newcommand{\condStd}{\textsc{KP-Standard}}
\newcommand{\condSal}{\textsc{KP-Salient}}
\newcommand{\condExp}{\textsc{Expert}}
\newcommand{\pp}{\,pp}
\newcommand{\PVR}{Keypoint Hand Visual Representation}
\newcommand{\task}{mug-in-box}
\newcommand{\Task}{Mug-in-Box}
\newcommand{\studyI}{Study~1}
\newcommand{\studyII}{Study~2}
\newcommand{\studyIII}{Study~3}
\newcommand{\novice}{novice}
\newcommand{\expert}{expert}
\newcommand{\pointConds}{keypoint-based condition}
\newcommand{\realShape}{real shape}
\newcommand{\RealShape}{Real shape}
\newcommand{\spc}{scene point cloud}
\newcommand{\Spc}{Scene point cloud}
\newcommand{\prop}{proprioceptive state}
\newcommand{\objPt}{object keypoint}
\newcommand{\ObjPt}{Object keypoint}
\newcommand{\handPt}{hand keypoint}
\newcommand{\collCond}{collection condition}
\newcommand{\obsRep}{observation representation}
\newcommand{\consist}{inter-trial consistency}
\newcommand{\Pset}{\mathcal{P}}
\newcommand{\piBase}{\pi_{\emptyset}}
\newcommand{\piPts}{\pi_{\mathcal{P}}}
\newcommand{\Lpos}{L_{\mathrm{pos}}}
\newcommand{\Lrot}{L_{\mathrm{rot}}}
\newcommand{\Lfin}{L_{\mathrm{fin}}}
\newcommand{\Jpos}{J_{\mathrm{pos}}}
\newcommand{\Jfin}{J_{\mathrm{fin}}}

\title{\LARGE \bf
  What Novice Operators See Changes What the Policy Learns:\\
  A User Study of Keypoint Hand Visual Representation\\
  for Teleoperated Demonstration
}

\author{Amane Yamaguchi$^{1*,2}$, Shohei Kamiguchi$^{1*}$, Masashi Hamaya$^{1}$,\\
  Shigeo Yoshida$^{1}$, Takuji Narumi$^{2}$, and Mai Nishimura$^{1}$%
  \thanks{This work has been submitted to the IEEE for possible publication. Copyright may be transferred without notice, after which this version may no longer be accessible.}%
  \thanks{\raggedright $^{1}$OMRON SINIC X Corporation, Tokyo, Japan.\newline
    \{shohei.kamiguchi, masashi.hamaya, shigeo.yoshida, mai.nishimura\}@sinicx.com}%
  \thanks{\raggedright $^{2}$The University of Tokyo, Tokyo, Japan.\newline
    \{yamaama, narumi\}@cyber.t.u-tokyo.ac.jp}%
  \thanks{$^{*}$Work done during internships at OMRON SINIC X.}%
}

\makeatletter
\@ifundefined{@IEEEtablestring}{}{%
  \long\def\@makecaption#1#2{%
    \ifx\@captype\@IEEEtablestring
      \begin{center}{\footnotesize #1}\\[2pt]%
        \parbox{\hsize}{\footnotesize #2}%
      \end{center}%
      \@IEEEtablecaptionsepspace
    \else
      \@IEEEfigurecaptionsepspace
      \setbox\@tempboxa\hbox{\footnotesize #1.~~ #2}%
      \ifdim \wd\@tempboxa >\hsize
        \setbox\@tempboxa\hbox{\footnotesize #1.~~ }%
        \parbox[t]{\hsize}{\footnotesize \noindent\unhbox\@tempboxa#2}%
      \else
        \hbox to\hsize{\footnotesize\box\@tempboxa\hfil}%
      \fi
    \fi}%
}
\makeatother

\makeatletter
\let\@oldmaketitle\@maketitle
\renewcommand{\@maketitle}{\@oldmaketitle
  \begin{center}
    \includegraphics[width=\textwidth]{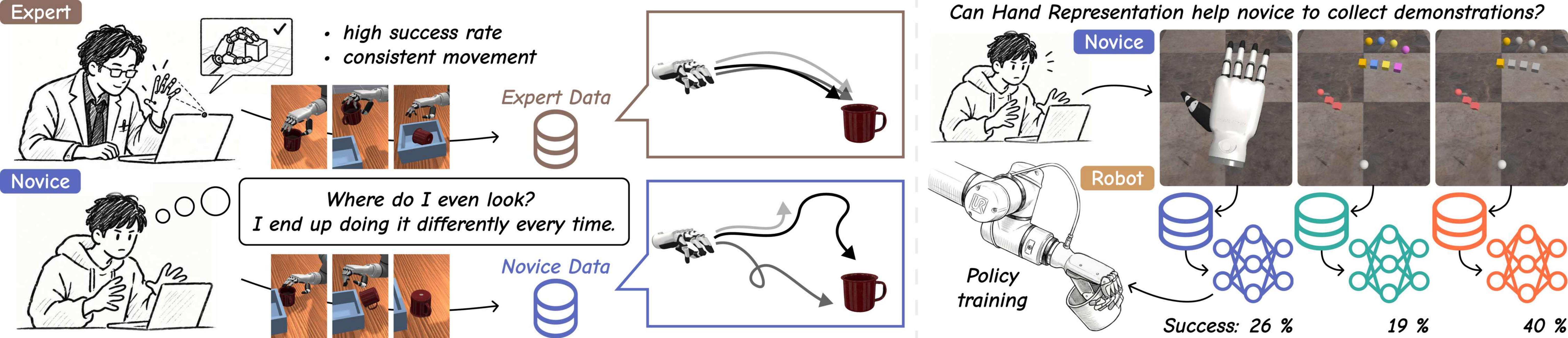}
  \end{center}
  \setcounter{figure}{0}\refstepcounter{figure}\label{fig:teaser}
  {\footnotesize Fig.~\thefigure.~~%
  Experts and \novice{}s perform the same \task{} task by teleoperation.
  Left: \novice{}s succeed less often than \expert{}s and repeat the task less consistently from trial to trial.
  Right: we replace the robot hand shown to the operator with sparse keypoints at its fingertips and joints, and give the same keypoints to the policy as observations.
  We collect demonstrations under three renderings of the hand and train a policy on each, and the success rates shown are those of the keypoint-observing policy within 200 steps.}\par
  \vspace{0.5em}}
\makeatother

\begin{document}

\maketitle

\bstctlcite{BSTcontrol}

\begin{abstract}
Imitation learning increasingly relies on \novice{} operators, whose demonstrations fall short of \expert{}s' in number and \consist{}.
Operators act on what they are shown, yet how the rendering of the robot affects their demonstrations, and the policies trained on them, has not been examined.
We study this question with \method{}, a teleoperation interface that renders a multi-fingered robot hand as sparse keypoints at its fingertips and joints, the same keypoints a policy can observe.
In a simulated \task{} task, 18 \novice{}s and 4 \expert{}s collected 2,200 trials under three renderings: the \realShape{} of the hand, all keypoints emphasized, or the thumb and index finger alone emphasized.
The \novice{}s shown the thumb and index finger emphasized collected 441 successful demonstrations from 600 trials, against 320 for those shown the \realShape{}.
We then trained diffusion policies on these demonstrations, replacing the \spc{} and \prop{} they observe with the hand keypoints and two \objPt{}s.
Within 400 steps these demonstrations went from the weakest of the three \novice{} conditions under the point-cloud policy (46.3\%) to the strongest under the keypoint policy (57.3\%), a gain of $+11.0$ percentage points in all five seeds.
Those collected with all keypoints emphasized changed by at most $+0.7$ percentage points, and in no more than three seeds.
The value of a demonstration therefore depends on the rendering shown during collection and on what the policy observes together.
A shared representation can pay off on both sides, in the number of successful demonstrations and in the success rate of the policy trained on them.
Project page: \projecturl.
\end{abstract}

\section{Introduction}
\label{sec:intro}

In imitation learning for multi-fingered robot hands, the quality of the demonstrations collected by \novice{} operators limits the performance of the policy.
As policies and datasets scale up~\cite{OXE2024,Khazatsky2024DROID}, imitation learning is expected to rely increasingly on \novice{}s who have no knowledge of policy learning or robot kinematics.
Multi-fingered hands make this especially visible: teleoperation constrains observation and action more than one's own body~\cite{Pei2024ArmRobot}, and the number of possible postures is vast, so consistent control takes practice.

What makes a demonstration good is \consist{}, and it has to hold with respect to what the policy observes rather than to the operator alone.
Sakr et al. define the quality of demonstration data by how repeatably the operator performs the same task~\cite{sawhney2024consistency}, and a policy can only learn from repetitions it can tell apart in its own observation~\cite{Belkhale2023DataQuality}.
Operators act on what they are shown, so the rendering of the robot decides which cues they rely on.
When those cues are absent from the policy's observation, demonstrations that were consistent for the operator can look inconsistent to the policy, an effect related to the imitation gap~\cite{weihs2021bridging}.
How the rendering shown to the operator affects the demonstrations and the policies trained on them has nevertheless not been examined.

This paper investigates this question with \emph{\method{}}, a teleoperation interface that renders the robot hand to the operator as sparse keypoints at the fingertips and at the joints that bend each finger (Fig.~\ref{fig:teaser}), a \pvr{}.
Unlike the \realShape{}, which occupies much of the view, the keypoints remove occlusion and limit the operator's cues to these parts, so we expect the task to become easier and the number and \consist{} of demonstrations to increase.
We compare three renderings of the hand: its \realShape{} (\condBase{}), all 11 keypoints emphasized (\condStd{}), and only the five keypoints on the thumb and index finger emphasized (\condSal{}).
In a user study, each \novice{} collected demonstrations under one of the three renderings, and the \expert{}s under the \realShape{} as a reference.
We train the same diffusion policy under two \obsRep{}s and compare them.
DP3~\cite{Ze2024DP3} observes the dense \spc{} with the \prop{}.
\emph{\kdp{}} replaces both with a sparse set of 13 points, the same \handPt{}s and two \objPt{}s on the symmetry axis of the object.

We ask three questions.
RQ1: does the \pvr{} bring the number and \consist{} of \novice{} demonstrations closer to the \expert{} level?
RQ2: how does the \collCond{} change the performance of policies trained on these demonstrations?
RQ3: does the effect of switching the policy's observation from the \spc{} and \prop{} to the hand and \objPt{}s depend on how the demonstrations were collected?
Our contributions are as follows.
\begin{itemize}
  \item We built \method{}, a teleoperation interface that varies the hand rendering shown to the operator while giving the same keypoints to the policy.
  \item We show that emphasizing only the thumb and index finger yielded the most successful \novice{} demonstrations, 441 of 600 trials against 320 for the \realShape{}, and the most consistent on four of the five kinematic metrics (\studyI{}).
  \item We find that in the fraction of episodes that place the mug within 400 steps, these demonstrations were the weakest of the \novice{} conditions under DP3 and the strongest under \kdp{}, so the rendering at collection and what the policy observes are worth choosing together (\studyII{}).
\end{itemize}

\section{Related Work}
\label{sec:related}

One approach to obtain better demonstrations is to make the robot easier for the operator to control.
Retargeting maps human hand motion to the robot so that the input stays accurate~\cite{qin2023anyteleop,handa2020dexpilot}.
Other methods adjust the camera viewpoint~\cite{jia2024viewpoint}.
Diminished reality, surveyed by Mori et al.~\cite{mori2017diminished}, removes the robot from the view so that the operator can see the workspace behind it.

The other approach leaves the operator alone and works on the data after collection.
Data curation discards low-quality demonstrations to improve policy performance~\cite{Deng2025Filtering,Chen2025Curating,Hejna2025DataCuration}.
Keypoint annotation reduces the dimensions the policy must learn~\cite{manuelli2019kpam,Corl2025PointPolicy}.
Curation, however, increases the collection time per retained demonstration, and keypoints require defining what matters for the robot and the object.
The first approach improves control without asking what that does to the data.
The second improves the data without touching how it was produced.
Neither treats the operator's view of the robot as something that shapes the demonstrations and, through them, the policy.

Closest to this question is work that aligns what the operator sees with what the policy observes.
Open-TeleVision~\cite{cheng2024opentelevision} streams the policy's camera images to the operator.
Other methods re-render the robot in collected data to match the policy's input~\cite{chen2024mirage,xu2025dexumi}.
Both show that alignment helps, but they align a stream or data that already exists.
The rendering is never designed at collection time for what it will do to the demonstrations.

Designing that rendering raises the question of how far the robot's appearance has to be retained.
Humans read biological motion from a few moving points~\cite{johansson1973visual}.
Studies of VR avatar embodiment report that the sense of embodiment depends on more than visual fidelity~\cite{fribourg2020controlling}.
Operators have even embodied a swarm of discrete robots and acted through it~\cite{ichihashi2024swarm}.
A hand reduced to keypoints at the fingertips and joints may therefore remain controllable.
Because those keypoints come from forward kinematics, they can also be given to the policy.

\section{Teleoperation System with a \PVR{}}
\label{sec:method}

\subsection{System Overview}
\label{sec:overview}

Through hand tracking, the operator controls a robot consisting of a multi-fingered hand and an arm.
We map finger motion to the robot with the DexPilot objective~\cite{handa2020dexpilot} in \texttt{dex-retargeting}, the public implementation of AnyTeleop~\cite{qin2023anyteleop}.
The hand-to-robot scale is 1.15 and no temporal filter is applied.
The task space is shown to the operator on a 2D display through a fixed camera.
The arm does not contribute directly to the task, so it is rendered transparent.
Both the transparency of the arm and the hand rendering described below apply only to the operator's view.
The depth camera that produces the policy's observation has the same pose as the display camera and sees the same scene in all conditions, so the \spc{} always contains the \realShape{} of the hand.

\subsection{Problem Formulation and Notation}
\label{sec:formulation}

We collect a demonstration set $\mathcal{D} = \{(o_t, a_t)\}_{t=1}^{T}$ by teleoperation, logging at each time step $t$ the observation $o_t$ and the action $a_t$ commanded by the operator.
The observation $o_t = (o_t^{\mathrm{pcd}}, o_t^{\mathrm{joint}})$ combines an exteroceptive and a proprioceptive component.
The \spc{} $o_t^{\mathrm{pcd}} \in \mathbb{R}^{N \times 3}$ is cropped from a depth camera and downsampled by farthest-point sampling.
The \prop{} decomposes into three components:
\begin{equation}
  o_t^{\mathrm{joint}} = (x_t,\; q_t,\; \phi_t), \quad x_t \in \mathbb{R}^3,\; q_t \in S^3,\; \phi_t \in \mathbb{R}^{d},
\end{equation}
where $x_t$ is the three-dimensional wrist position, $q_t$ is the unit quaternion of the wrist orientation, and $\phi_t$ holds the joint angles of the $d$ actuated joints of the hand, with $d$ a constant of the hardware.
All three are measured in the simulator.
The action $a_t \in \mathbb{R}^3 \times S^3 \times \mathbb{R}^{d}$ is a target in the same parameterization, so observations and actions lie in the same space.
Closed-loop inverse kinematics in Pinocchio turns the target wrist pose into arm joint targets, and the finger targets are sent as joint commands.
This conversion is the same during collection and during execution.

\subsection{Design of the \PVR{}}
\label{sec:design}

\begin{figure}[t]
  \centering
  \includegraphics[width=0.32\columnwidth]{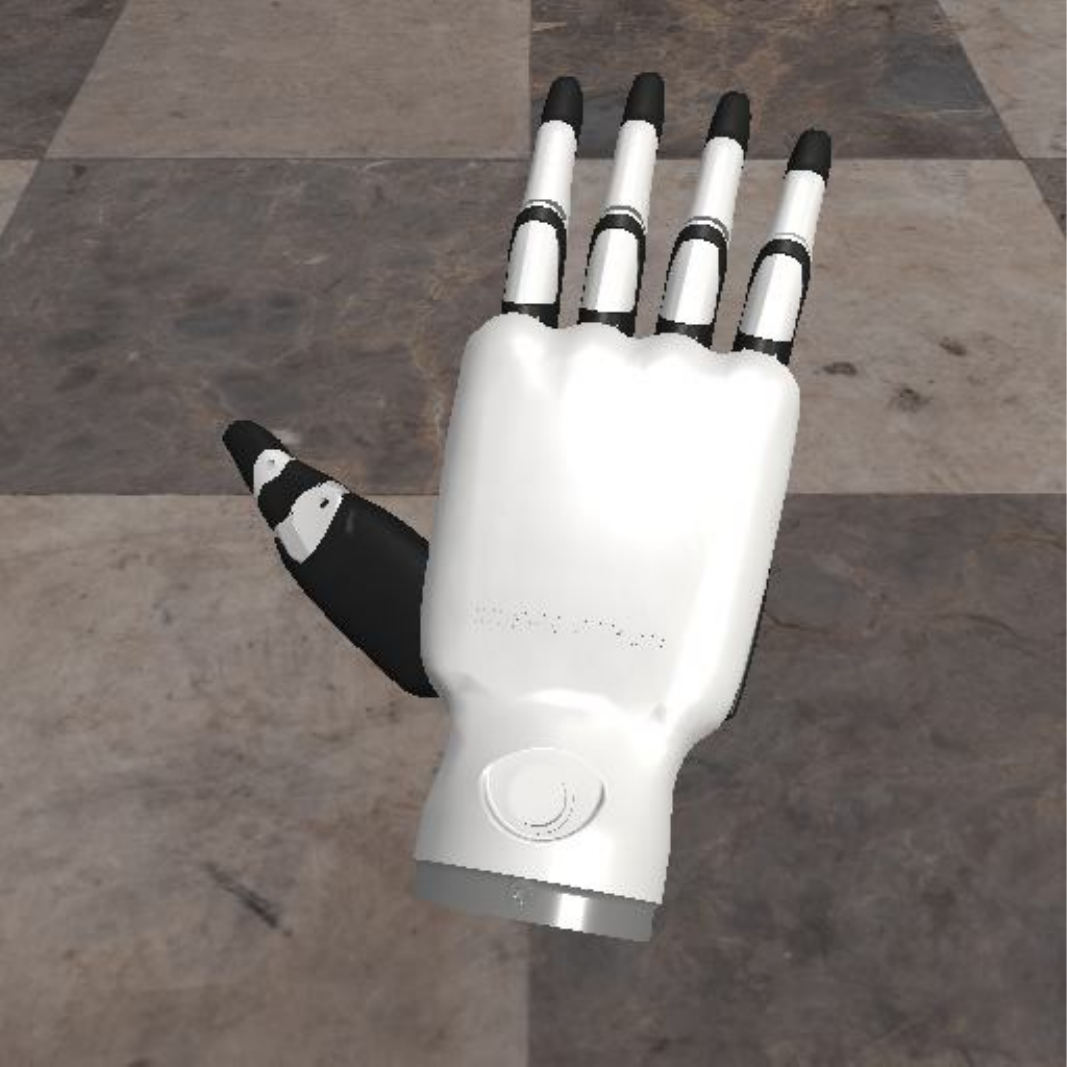}\hfill
  \includegraphics[width=0.32\columnwidth]{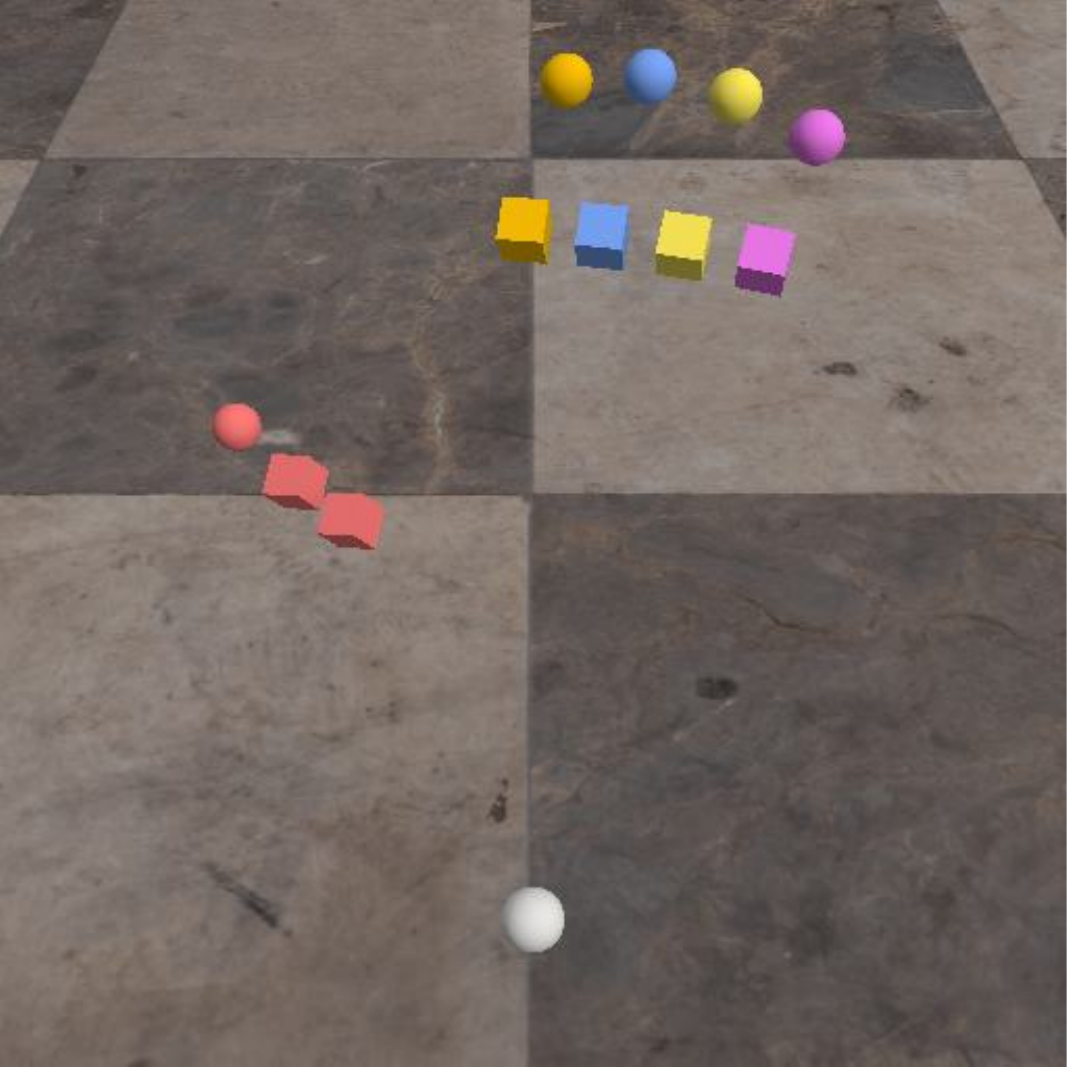}\hfill
  \includegraphics[width=0.32\columnwidth]{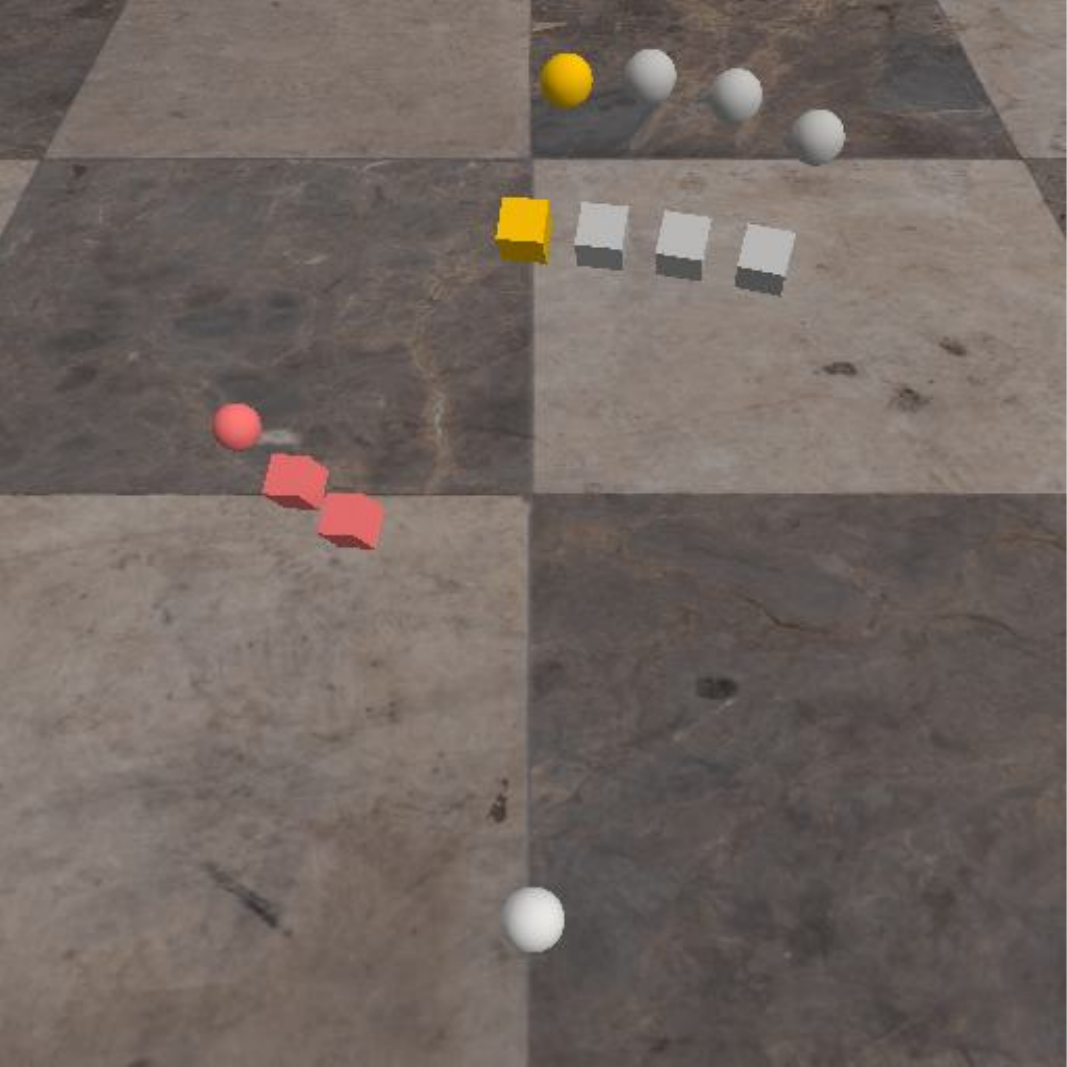}
  \caption{Robot hand presented to the operator: \realShape{} (left, \condBase{}), all keypoints emphasized (center, \condStd{}), and only those of the thumb and index finger emphasized (right, \condSal{}).
  Fingertip keypoints are spheres (radius 0.007\,m) and intermediate joint keypoints cubes (edge 0.012\,m). Emphasized keypoints have a per-finger color that brightens on contact. The others and the wrist keypoint, which the policy does not receive, are gray.}
  \label{fig:conditions}
\end{figure}

\method{} replaces the \realShape{} presented to the operator with sparse keypoints $\Pset{} \in \mathbb{R}^{K \times 3}$, placed at the fingertip of each finger and at the joints that bend it (Fig.~\ref{fig:conditions}).
Their number $K$ is set by the kinematics of the hand, and every keypoint position follows from $o_t^{\mathrm{joint}}$ by forward kinematics.
Each keypoint is therefore available whether or not the camera sees the finger it belongs to.
The \realShape{} occludes itself, so to see the region behind the hand the operator either moves the hand aside or predicts what it hides.
The keypoints remove this occlusion and become the only parts of the hand the operator can refer to.

Of the five design axes of the \pvr{}, the two \pointConds{}s in our experiments differ only in which keypoints are emphasized, either all of them or those of the thumb and index finger alone.
The emphasis also decides which keypoints carry the contact cue.
Emphasized keypoints are given a per-finger color that brightens toward white by 35\% while the keypoint is within 0.08\,m of the object, and the others are gray and do not respond to contact.
The remaining three axes (the placement and number of the keypoints, their shape and size, and their color) were fixed in a preliminary study that compared renderings of different keypoint densities.
Fingertip keypoints are drawn in a shape closer to the \realShape{} than intermediate joint keypoints, and a gray wrist keypoint outside $\Pset{}$ is added.

\subsection{Policy Training}
\label{sec:policy}

\begin{figure}[t]
  \centering
  \includegraphics[width=\columnwidth]{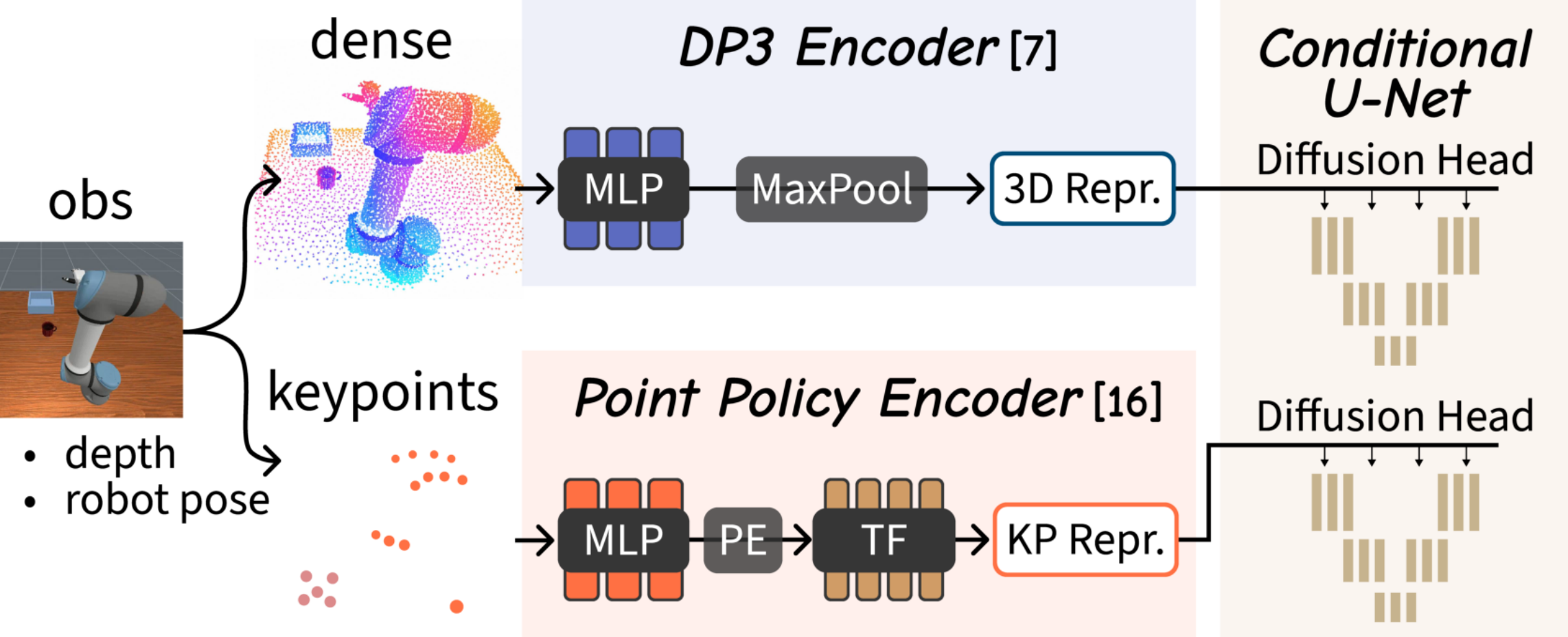}
  \caption{Observation encoding of the two policies.
  Top: DP3 ($\piBase{}$) uses the DP3 encoder~\cite{Ze2024DP3}, which encodes the dense \spc{} with a PointNet-based MLP of widths 64, 128, 256 and max-pooling, projected to a 64-dimensional feature and concatenated with a 64-dimensional state feature.
  Bottom: \kdp{} ($\piPts{}$) uses the Point Policy encoder~\cite{Corl2025PointPolicy}, which embeds each hand and \objPt{} with a shared layer of width 512 projected to 256, adds a learnable per-keypoint positional embedding (PE), and passes each as one token to a four-layer, two-head Transformer, so the correspondence between tokens and robot parts is preserved.
  Both condition a diffusion head of the same architecture and share the training procedure.}
  \label{fig:policy}
\end{figure}

To ask how the value of a demonstration depends on what the policy observes, we train the same diffusion policy under two encodings (Fig.~\ref{fig:policy}): a dense order-free encoding of the \spc{} and a sparse order-preserving encoding of the keypoints.
The keypoint positions are recorded with every demonstration, whether or not they were shown to the operator, so both encodings are trained on the same demonstrations.

DP3 ($\piBase{}$)~\cite{Ze2024DP3} encodes the \spc{} into a compact 3D feature with a PointNet-based encoder and the \prop{} into a second feature with a lightweight MLP.
The conditioning vector is fed to a ConditionalUNet1D denoiser, which is trained with DDPM and generates action chunks with DDIM inference.

\kdp{} ($\piPts{}$) replaces both inputs of DP3 with the positions of the hand keypoints and of two \objPt{}s.
Following Point Policy~\cite{Corl2025PointPolicy}, it embeds the history of each keypoint over the observation horizon with a shared MLP, adds a learnable per-keypoint positional embedding, and passes each of the 13 keypoints to a Transformer as one token.
The output tokens are concatenated in the order of the keypoints, so unlike the permutation-invariant aggregation of PointNet, the correspondence between tokens and robot parts is preserved.
\kdp{} receives the keypoint positions but not their colors or brightness, so the choice of which keypoints are emphasized acts only on the operator.
The action head and the training procedure are shared with DP3.
Each encoder is used as published rather than matched to the other, so the two differ simultaneously in the replaced input, the per-keypoint embedding, the token-wise Transformer, and the added \objPt{}s.

The two \objPt{}s lie on the rotational symmetry axis of the object, at the centers of its bottom and its opening.
They represent the position and the axis of the object, but not the rotation about that axis.
Before grasping, each side of a bounding box of the object is shrunk about its center to half its length.
The keypoints are then placed on the vertical axis through the mean of the depth points inside this box, leaving out the points within 0.15\,m of the wrist.
This estimate is corrected by an offset calibrated from demonstrations.
A grasp is detected from the tracking error and the opening of the fingers, after which the keypoints are propagated rigidly with the wrist pose.

\section{Experiments}
\label{sec:experiments}

We conducted three studies.
\studyI{} is a user study on how the hand rendering shown to the operator affects the number and \consist{} of \novice{} demonstrations (RQ1).
\studyII{} measures the performance of policies trained on those demonstrations (RQ2 and RQ3).
\studyIII{} is a preliminary check of whether the trained policies operate on a physical robot.

The robot consists of an Inspire Hand (five fingers, underactuated, $d=6$ actuated joints) and a UR5e arm, so $o_t^{\mathrm{joint}}$ and $a_t$ are 13-dimensional.
Control runs at 20\,Hz, and we call one control period a step.
The thumb bends two joints independently and each other finger bends one, so $\Pset{}$ has 11 keypoints: three on the thumb (intermediate joint, distal joint, and tip) and two on each of the other four fingers (intermediate joint and tip).
Studies~1 and~2 used ManiSkill~\cite{tao2025maniskill3} on the SAPIEN physics engine~\cite{xiang2020sapien} at 200\,Hz, and \studyIII{} used a physical robot of the same configuration.
The operator's hand was tracked by a Leap Motion 2 (Ultraleap).

\subsection{\Task{} Task}
\label{sec:task}

All three studies use the same task of placing a mug on a table into a box placed farther back.
The mug is 025\_mug from the YCB dataset.
The box has inner dimensions of $200 \times 200$\,mm and 80\,mm high walls, and is fixed 0.20\,m forward of the table center.
Each trial starts with the mug upright at a uniformly random position within $[-0.14, 0]$\,m forward and $[-0.07, 0.07]$\,m lateral of the table center, at a random orientation about the vertical axis.
A trial ends in success when the center of the mug is inside the box, above its floor and below the top of its walls, so a mug still held above the box does not count.
It ends in failure when the mug drops 0.2\,m below the table, leaves $\pm 0.5$\,m from the table center, or moves more than 0.95\,m from the robot base.
A trial also ends in failure when the time limit given in each study expires.

\subsection{\studyI{}: User Study on Demonstration Consistency}
\label{sec:study1}

\studyI{} answers RQ1, whether the \pvr{} brings the number and \consist{} of \novice{} demonstrations closer to the \expert{} level.
It used a between-participants design with the visual representation as the single factor, to avoid carry-over of learning between conditions.
\condBase{} keeps the \realShape{} of the hand, and both \pointConds{}s replace it with the 11 keypoints (Table~\ref{tab:conditions}).
\condStd{} emphasizes all 11, whereas \condSal{} emphasizes only the five on the thumb and index finger and draws the remaining six in gray.
Because the \condBase{} hand does not brighten on contact, the two \pointConds{}s differ from \condBase{} in both occlusion and the visual contact cue.

\begin{table}[t]
  \centering
  \caption{Visual representation conditions assigned to \novice{}s in \studyI{}. The \expert{}s used \condBase{}.}
  \label{tab:conditions}
  \setlength{\tabcolsep}{4pt}
  \begin{tabular}{@{}llll@{}}
    \toprule
    Condition & Hand rendering & Emphasized keypoints & Contact cue \\
    \midrule
    \condBase{} & \RealShape{} & --- & None \\
    \condStd{} & 11 keypoints & All fingers (11) & Brightness \\
    \condSal{} & 11 keypoints & Thumb and index (5) & Brightness \\
    \bottomrule
  \end{tabular}
\end{table}

We predicted that, in the two \pointConds{}s, reduced occlusion would raise the success rate during collection without increasing workload relative to \condBase{}.
We further predicted that \condSal{} would bring \consist{} closer to the \expert{} level.
The grasp in the \task{} task is carried by the thumb and index finger, so emphasizing only their keypoints leaves the operator fewer parts to read and concentrates attention on the ones that decide the outcome.

Nineteen \novice{}s were randomly assigned to the three conditions, seven to \condStd{} and six to each of the others.
None of them had operated a robot routinely, and all reported no robot operation skill.
As a reference, four \expert{}s took part in \condBase{} only (\condExp{}), one of them the first author.
They had research experience in robot imitation learning or operator experience in robot competitions, self-reported and confirmed orally.
The \expert{} level is therefore a reference measured under the \realShape{}.
Adults of any gender were recruited through social media and a recruitment service, and received 2,000 JPY (approximately 14 USD) for a session of up to 90 minutes.
The experiment was approved by the ethics review board of the authors' institution.

After giving consent, each participant practiced under their assigned rendering until at least 10 minutes had elapsed and 20 trials had succeeded, with each practice trial limited to 752 steps (37.6\,s).
Before data collection we set a limit of 45 minutes on practice, and a participant who reached it was withdrawn at that point.
This happened to one \condStd{} participant before any main trial, leaving 18 \novice{}s (six per condition).
Each participant then performed 100 main trials in two blocks of 50, with each trial limited to 376 steps (18.8\,s).
The experimenter restarted a trial when hand tracking dropped out and the arm consequently passed through a kinematic singularity, which moves the hand in a way the operator did not command.
Such trials were not counted.
Finally, participants answered the NASA Task Load Index (NASA-TLX)~\cite{hart1988nasatlx}, a questionnaire on the workload they experienced during the main trials.

Besides the number of successful demonstrations and the success rate during collection, we measured five kinematic metrics in the three spaces of the decomposition $o_t^{\mathrm{joint}} = (x_t, q_t, \phi_t)$ (Sec.~\ref{sec:formulation}).
Path length and jerk are standard measures of trajectory quality~\cite{Flash1985MinimumJerk} that recent benchmarks compute per space~\cite{Wang2025RoboEval}, so we take path length for all three components and jerk for the two Euclidean ones:
\begin{align}
  \Lpos{} &= \sum_t \lVert x_{t+1}-x_t \rVert, &
  \Lrot{} &= \sum_t d_{\mathrm{geo}}(q_t, q_{t+1}), \nonumber\\
  \Lfin{} &= \sum_t \lVert \phi_{t+1}-\phi_t \rVert, \nonumber\\
  \Jpos{} &= \frac{1}{T}\sum_t \lVert x^{(3)}_t \rVert, &
  \Jfin{} &= \frac{1}{T}\sum_t \lVert \phi^{(3)}_t \rVert,
\end{align}
where $d_{\mathrm{geo}}$ is the geodesic angle between unit quaternions, $T$ is the number of steps, and $x^{(3)}_t$ and $\phi^{(3)}_t$ are third derivatives computed by finite differences.

Following Sakr et al.~\cite{sawhney2024consistency}, we quantify the \consist{} of each metric by its within-participant coefficient of variation (CV), where a smaller CV means higher consistency.
The CV is taken over the successful main trials of one participant and then averaged over the participants of a condition ($n=6$ per condition, $n=4$ for \expert{}s).
With this sample size, \studyI{} results are reported descriptively without statistical tests.

\subsubsection{Results}

\begin{figure}[t]
  \centering
  \includegraphics[width=\columnwidth]{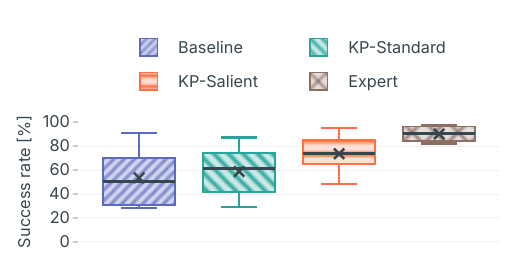}
  \caption{Success rate during collection, one value per participant ($n=6/6/6/4$).
  Boxes show the median as a dark line and the mean as a cross, and the hatch pattern identifies the condition.}
  \label{fig:success}
\end{figure}

\textit{Number and success rate.}
\condSal{} yielded the most successful demonstrations, 441 of 600 trials, compared with 320 for \condBase{} and 352 for \condStd{} (360 of 400 for \condExp{}).
The success rate during collection, computed by dividing each participant's successful trials by the 100 main trials, was $53.3 \pm 26.1$\% for \condBase{}, $58.7 \pm 22.7$\% for \condStd{}, $73.5 \pm 17.1$\% for \condSal{}, and $90.0 \pm 7.6$\% for \condExp{} (mean $\pm$ SD over participants, Fig.~\ref{fig:success}).
Excluding the first author, the most successful of the four \expert{}s, lowers \condExp{} to 87.7\% and leaves the ordering of the success rates unchanged.

\begin{figure*}[!tb]
  \centering
  \includegraphics[width=\textwidth]{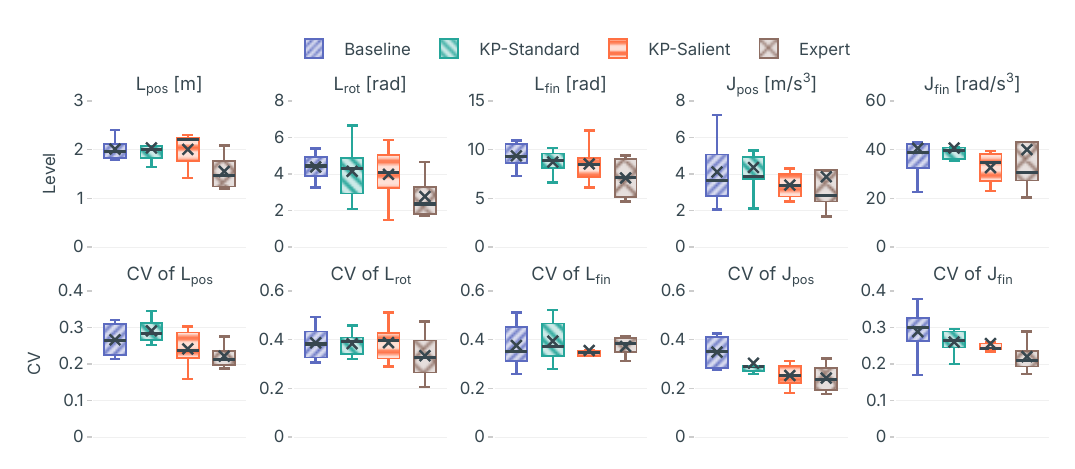}
  \caption{Levels of the five main metrics (top) and their within-participant coefficients of variation (bottom).
  Each box summarizes one value per participant ($n=6/6/6/4$), with the median as a dark line and the mean as a cross. The hatch pattern identifies the condition.}
  \label{fig:consistency}
\end{figure*}

\begin{figure*}[!tb]
  \centering
  \includegraphics[width=\textwidth]{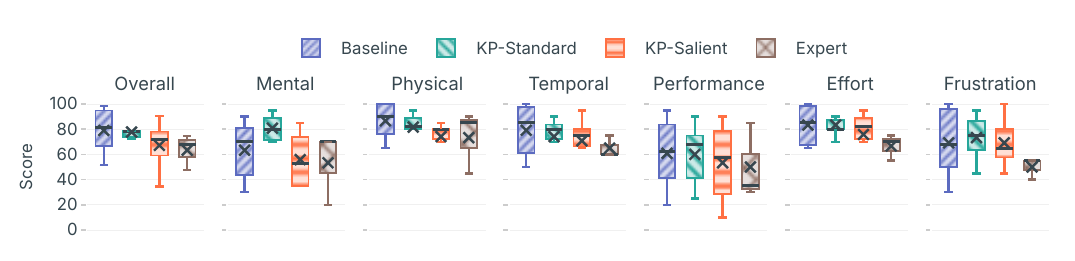}
  \caption{Overall NASA-TLX score (weighted) and six subscales (Mental, Physical, and Temporal Demand, Performance, Effort, Frustration) per condition.
  The first author, one of the \expert{}s, knew the purpose of the experiment and did not answer ($n=6/6/6/3$).
  Boxes show the median as a dark line and the mean as a cross. Hatch patterns identify the condition.}
  \label{fig:tlx}
\end{figure*}

\textit{Kinematic metrics.}
Among the three \novice{} conditions, $\Lrot{}$ and $\Lfin{}$ were shortest in \condSal{}, whereas $\Lpos{}$ differed by at most 0.03\,m (Fig.~\ref{fig:consistency}, top).
Both $\Lrot{}$ and $\Lfin{}$ were shorter in \condExp{} than in any \novice{} condition.
For \condBase{}/\condStd{}/\condSal{}/\condExp{}, $\Lrot{}$ was 4.38/4.14/3.98/2.76\,rad, $\Lfin{}$ was 9.35/8.72/8.53/7.09\,rad, and $\Lpos{}$ was 2.01/2.03/2.00/1.55\,m.

\textit{Inter-trial consistency.}
For both jerk metrics, the within-participant CVs of the two \pointConds{}s were smaller than in \condBase{} (Fig.~\ref{fig:consistency}, bottom).
For \condBase{}/\condStd{}/\condSal{}/\condExp{}, $\Jpos{}$ was 0.348/0.303/0.252/0.242 and $\Jfin{}$ 0.288/0.260/0.256/0.220.
Excluding the first author raises the \condExp{} CV of $\Jpos{}$ from 0.242 to 0.256, which no longer falls below \condSal{}, and leaves the other orderings unchanged.
For path length, the CVs of the two \pointConds{}s moved in opposite directions from \condBase{}.
For \condBase{}/\condStd{}/\condSal{}, $\Lpos{}$ was 0.266/0.290/0.241 and $\Lfin{}$ 0.376/0.393/0.355, while $\Lrot{}$ ranged only from 0.383 to 0.387.

\textit{Workload.}
Among the \novice{} conditions, the overall NASA-TLX score was lowest in \condSal{} (67.2, compared with 79.0 for \condBase{}, 77.8 for \condStd{}, and 63.3 for \condExp{}, Fig.~\ref{fig:tlx}).
Mental Demand separated the \novice{} conditions the most, from 80.8 in \condStd{} to 55.8 in \condSal{} (63.3 for \condBase{}), a spread of 25.0 points against 4.2--12.5 points for the other five subscales.
\condSal{} was also the lowest of the three \novice{} conditions on those five, tied with \condBase{} on Frustration.

\begin{figure}[t]
  \centering
  \includegraphics[width=\columnwidth]{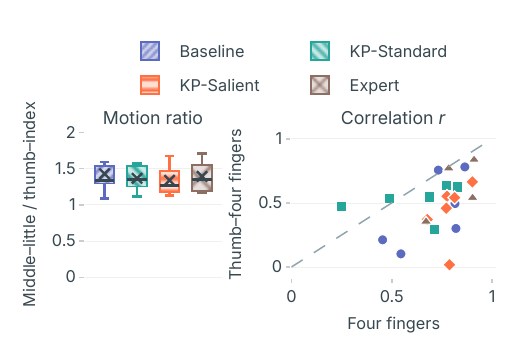}
  \caption{Post hoc analysis of finger joint targets in successful trials: motion ratio of the middle, ring, and little fingers to the thumb and index finger (left) and four-finger against thumb--four-finger correlation (right, one marker per participant, $n=6/6/6/4$, dashed line $y=x$).
  Boxes show the median as a dark line and the mean as a cross. Hatch patterns and marker shapes (circle \condBase{}, square \condStd{}, diamond \condSal{}, triangle \condExp{}) identify the condition.}
  \label{fig:finger}
\end{figure}

\textit{Finger usage.}
In a post hoc analysis of the finger joint targets (Fig.~\ref{fig:finger}), the middle, ring, and little fingers moved 1.34 to 1.43 times as much per step as the thumb and index finger in all four conditions.
The ratio was lowest in \condSal{}.
Their mean absolute change ranged from 0.0163 to 0.0191\,rad/step.
The mean pairwise Pearson correlation among the four fingers other than the thumb exceeded their correlation with the thumb for 19 of the 22 participants.

\subsection{\studyII{}: Policy Learning from the Collected Demonstrations}
\label{sec:study2}

\begin{figure*}[!tb]
  \centering
  \includegraphics[width=\textwidth]{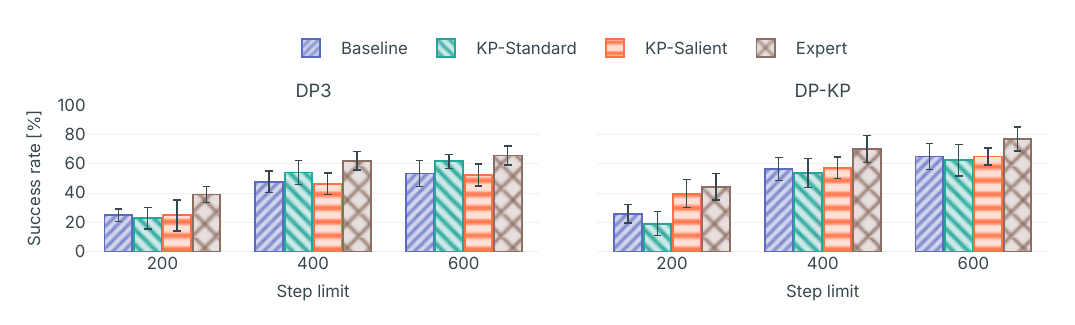}
  \caption{Simulated success rate within 200, 400, and 600 steps, one panel per \obsRep{} and one bar per \collCond{}. Bars show the mean over five seeds and error bars the seed-to-seed standard deviation, and the hatch pattern identifies the \collCond{}.}
  \label{fig:study2}
\end{figure*}

\studyII{} answers RQ2, how the \collCond{} changes the performance of policies trained on these demonstrations, and RQ3, whether the effect of switching the policy's observation depends on how the demonstrations were collected.
It measures the value of the demonstrations by the performance of policies trained on them.
The design has two factors, the \collCond{} (the three conditions of \studyI{} and \condExp{}) crossed with the \obsRep{} (DP3 or \kdp{}, Sec.~\ref{sec:policy}).
We predicted that the success rate would follow the \collCond{}s in the order \condBase{} $<$ \condStd{} $<$ \condSal{} under both \obsRep{}s, and that data collected with only the thumb and index finger emphasized would gain the most from \kdp{}.
Because the two encoders differ in architecture as well as in what they observe (Sec.~\ref{sec:policy}), we read the change from DP3 to \kdp{} only through how it differs across \collCond{}s.
It is not evidence that one encoder is better than the other.

For each condition and each of five seeds, every configuration was trained on 100 demonstrations sampled uniformly from the pooled successful trials of all participants.
Fixing this number separates the value of the demonstrations from the difference in their number found in \studyI{}.
Sampling from the pooled trials, 320/352/441/360 for \condBase{}/\condStd{}/\condSal{}/\condExp{}, weights each participant by the number of trials they succeeded in.
The network initialization was fixed, so the seeds vary only the sampled demonstrations.

Both policies were trained with the settings in Table~\ref{tab:settings} and the remaining hyperparameters of DP3.\footnote{100 diffusion steps for training and 10 for inference, an action chunk length of 16, an observation horizon of 2, and 8 executed steps per chunk.}
The weights were selected on 10\% of the training samples held out for validation.

\kdp{} never receives the ground-truth pose of the object, only \objPt{}s estimated from depth as in Sec.~\ref{sec:policy}.
The bounding box that localizes them was obtained by projecting the object mesh at its ground-truth pose, which stands in for an ideal detector without detection error.
The gains we report for \kdp{} over DP3 are therefore an upper bound.
The same idealization is applied in every \collCond{}, so it cannot account for the differences between them.

The policies differ in how quickly they place the mug as well as in whether they place it at all, so we report three step limits rather than one.
The initial mug placements are shared across the eight configurations, and each episode runs for up to 1000 steps.
We report the fraction completed within 200, 400, and 600 steps, and at the full budget, as the mean and standard deviation over the five seeds.
Differences between success rates are given in percentage points (pp).
For the change from DP3 to \kdp{}, we report paired $t$-tests over seeds, since the same 100 demonstrations are given to both policies for each condition and seed.
The \collCond{} has a single participant pool per condition, so we test neither that factor nor the interaction of the two factors.
We compare \collCond{}s by the paired differences themselves and by the number of seeds in which they hold, since one test reaching significance while another does not is weak ground for separating two conditions.

\subsubsection{Results}

\textit{Change from DP3 to \kdp{}.}
Switching from DP3 to \kdp{} changed the success rate at 200/400/600 steps by $+15.0$ ($p=0.020$)/$+11.0$ ($p=0.039$)/$+12.7$\pp{} ($p=0.015$) for the \condSal{} data and by $-3.7$/$-0.3$/$+0.7$\pp{} ($p \ge 0.61$) for the \condStd{} data (Fig.~\ref{fig:study2}).
For the \condBase{} data it changed by $+1.0$ ($p=0.73$)/$+8.7$ ($p=0.004$)/$+11.7$\pp{} ($p=0.024$), and for the \condExp{} data by $+5.3$ ($p=0.30$)/$+8.0$ ($p=0.19$)/$+11.3$\pp{} ($p=0.065$).
The change was positive in all five seeds at all three step limits for \condSal{}, in 1/3/2 of the five seeds for \condStd{} (per-seed changes from $-23.3$ to $+20.0$\pp{}), in 2/5/5 for \condBase{}, and in 3/4/5 for \condExp{}.
At 1000 steps the change was $+13.7$\pp{} ($p=0.005$) in five of five seeds for \condSal{} and $+1.7$\pp{} in three of five for \condStd{}.

\textit{Effect of the \collCond{}.}
With DP3, the \condSal{} data ranked lowest of the three \novice{} conditions at 400 and 600 steps (46.3\% and 52.3\%), $7.7$ and $9.3$\pp{} below \condStd{}.
With \kdp{}, they ranked highest at 200 and 400 steps (39.7\% and 57.3\%), above the next \novice{} condition by 14.0 and 1.0\pp{}, and level with \condBase{} at 600 steps (65.0\%).
At 1000 steps the \condSal{} data were lowest of the \novice{} conditions with DP3 (56.3\% against 65.7\% for \condStd{}) and highest with \kdp{} (70.0\% against 69.7\% for \condBase{}).
The gap between \condSal{} and \condExp{} was 14.3/15.7/13.3\pp{} at 200/400/600 steps with DP3, compared with 4.7/12.7/12.0\pp{} with \kdp{}.

\begin{table}[t]
  \centering
  \caption{Policy training and evaluation settings in \studyII{}.}
  \label{tab:settings}
  \setlength{\tabcolsep}{4pt}
  \begin{tabular}{@{}ll@{}}
    \toprule
    Training data & 100 demonstrations per condition and seed \\
    Seeds & 5 (sampled demonstrations) \\
    \Spc{} points $N$ & 512 \\
    Epochs, batch size & 500, 128 \\
    Optimizer & AdamW, $10^{-4}$, weight decay $10^{-6}$, \\
      & 5 warmup epochs then cosine, clip 1.0 \\
    Evaluated weights & EMA, lowest validation loss \\
    \ObjPt{}s & depth inside an ideal-detector box \\
    Camera & $[-0.4,0,0.4]$\,m to $[-0.1,0,0.1]$\,m, \\
      & $45^\circ$ vertical, $224 \times 224$ px \\
    Evaluation & 60 episodes per seed, up to 1000 steps \\
    \bottomrule
  \end{tabular}
\end{table}

\subsection{\studyIII{}: Preliminary Study on a Physical Robot}
\label{sec:study3}

We transferred the trained policies to a physical robot (UR5e and Inspire Hand).
For each of the eight configurations, one policy was pre-trained for 500 epochs on all successful demonstrations of its \collCond{}, then fine-tuned for 100 epochs on five demonstrations collected on the physical robot.
The \spc{} was downsampled to $N=256$ to match the physical depth sensor, and the \objPt{}s were obtained by fitting a cylinder to the depth points instead of from a detector.
Each policy was evaluated in 10 trials, and under DP3 those trained on the \condBase{}/\condStd{}/\condSal{}/\condExp{} demonstrations succeeded in 5/6/7/6 of them.
The video on the project page\footnote{\projecturl} shows the resulting behavior.

\section{Discussion}
\label{sec:discussion}

RQ1 is answered in part.
In jerk, the \pvr{} brought the \consist{} of \novice{} demonstrations closer to the \expert{} level under both renderings.
In path length it did so only under \condSal{}, and there only in $\Lpos{}$ and $\Lfin{}$.
\condSal{} also yielded 441 successful demonstrations of 600 against 320 for \condBase{}, whereas \condStd{} was less consistent than \condBase{} in two of the three path lengths (Sec.~\ref{sec:study1}).
Under the \pvr{} the rotation and finger paths were about 5--9\% shorter than under the \realShape{}, a difference that lies within the between-participant spread.
Two explanations fit it equally well.
One is the removal of the self-occlusion that forces the operator to move the hand aside, or to predict, in order to see what lies behind it~\cite{mori2017diminished}.
The other is the visual contact cue.
A lower perceptual load separates the two conditions that both remove occlusion.
\condSal{} reached a higher success rate during collection than \condStd{} and also carried the lowest Mental Demand.
The middle, ring, and little fingers moved 1.34 to 1.43 times as much as the thumb and index finger in all four conditions (Sec.~\ref{sec:study1}), so the emphasis did not suppress the use of particular fingers.
The lower collection success rate of \condStd{} is more consistent with the load of its presentation than with a change in how the fingers were used.

What the policy observed reordered the \collCond{}s, which in \studyI{} ranked \condSal{} (441 successful demonstrations of 600), \condStd{} (352), and \condBase{} (320).
The \condSal{} data ranked lowest of the three \novice{} conditions under DP3 at 400 and 600 steps, and highest under \kdp{} at 200 and 400 steps.
The \condStd{} data moved the other way, highest under DP3 and lowest under \kdp{} at all three step limits (Sec.~\ref{sec:study2}).
Which \collCond{} produced the better policy therefore depended on the \obsRep{}, so RQ2 has no answer apart from RQ3.
Neither policy followed the predicted order \condBase{} $<$ \condStd{} $<$ \condSal{}.
The \condStd{} operators worked under the highest Mental Demand and collected fewer successful demonstrations than the \condSal{} operators (Sec.~\ref{sec:study1}), yet their demonstrations gave DP3 the highest success rate of the three \novice{} conditions.
Of the two conditions in which the operator saw the keypoints the policy receives, only the one that emphasized the thumb and index finger gained under \kdp{}.

These results imply that the rendering shown during collection and the \obsRep{} of the policy are worth choosing together.
The same \condSal{} demonstrations were the weakest of the \novice{} conditions under DP3 at 400 and 600 steps and the strongest under \kdp{} at 200 and 400 steps.
A representation shared by the operator and the policy can also pay off on both sides at once, but only when it is designed well.
For \condSal{} it paid off in the workload and the number of demonstrations during collection and in the success rate of \kdp{} afterwards.

\subsection{Limitations and Future Work}
\label{sec:limitations}

The placement and emphasis of the keypoints changed how the demonstrations came out, but which assignment yields the best demonstrations remains unknown.
We emphasized the thumb and index finger assuming they mainly perform grasping, yet most participants moved the four fingers other than the thumb together (Sec.~\ref{sec:study1}), a grouping that differs from this emphasis.
We used one hand on one task, and a different task, a hand with different kinematics, or a gripper will likely call for a different placement.
What is needed is a way to design the keypoints automatically, for instance from a few demonstrations.

The \pvr{} conveys contact only by brightness, so it cannot represent grip force or slip.
Extending it beyond grasping and transport to tasks dominated by force or touch will likely require combining it with force or haptic feedback.

The visual representation was manipulated only in simulation.
Presenting it to an operator of a physical robot requires removing the hand from the camera image without delay and overlaying the keypoints, for which real-time object removal with 3D Gaussian Splatting~\cite{Waldow2025DimSplat} is a promising means.

\section{Conclusion}
\label{sec:conclusion}

Using \method{}, an interface that renders the robot hand as sparse keypoints, we studied how the rendering shown to \novice{} operators affects their demonstrations and the policies trained on them.
Emphasizing the thumb and index finger yielded 441 successful demonstrations of 600 trials, against 320 for the \realShape{}.
Switching the policy's observation to the \handPt{}s and two \objPt{}s then raised their success rate in every seed at all three step limits.
It did so from 400 steps on for those collected with the \realShape{}, and not consistently for those collected with all keypoints emphasized.
These results suggest that the rendering shown during collection and what the policy observes should be designed together.

\section*{Acknowledgment}

This work was supported by JSPS KAKENHI Grant Number 25K22837 and JST CREST, Japan, Grant Number JPMJCR2552.
We used Claude Code (Anthropic) to translate and revise the manuscript from the authors' draft and to prototype the scripts for data analysis and plotting, and ChatGPT and Adobe Firefly to generate the illustrations of the operators in Fig.~\ref{fig:teaser}.
The authors verified all generated text, code, and numerical results against the data and take full responsibility for the content.

\bibliographystyle{IEEEtran}
\bibliography{root}

\end{document}